\documentclass[11pt,a4paper]{article}
\usepackage[hyperref]{emnlp2020}
\usepackage{times}
\usepackage{latexsym}
\usepackage{booktabs}
\usepackage{graphicx}
\usepackage{amsmath}
\usepackage{float}
\usepackage{graphics}
\usepackage{bm}
\usepackage{tikz-qtree}
\usepackage{dsfont}
\usepackage{textcomp}

\usepackage{microtype}
\usepackage{amsmath}
\usepackage{amssymb}
\usepackage{subcaption}
\usepackage{caption}
\usepackage[acronym]{glossaries}
\aclfinalcopy 

\newcommand{\architecturename}{OURS}

\newcommand{\fnversion}{FN1.5}

\newacronym{srl}{SRL}{Semantic Role Labeling}
\newacronym{gcn}{GCN}{Graph Convolutional Network}
\newacronym{fn}{FN15}{FrameNet 1.5}
\newacronym{cn}{CN05}{CoNLL-2005}

\title{Encoding Syntactic Constituency Paths for Frame-Semantic Parsing with Graph Convolutional Networks}

\author{Emanuele Bastianelli, Andrea Vanzo, and Oliver Lemon \\
	Interaction Lab, MACS, Heriot-Watt University, Edinburgh, UK \\
	{\tt \{e.bastianelli, a.vanzo, o.lemon\}@hw.ac.uk} \\
}

\date{}

\begin{document}
\maketitle
\begin{abstract}
We study the problem of integrating syntactic information from constituency trees into a neural model in Frame-semantic parsing sub-tasks, namely Target Identification (TI), Frame Identification (FI), and Semantic Role Labeling (SRL). We use a Graph Convolutional Network to learn specific representations of constituents, such that each constituent is profiled as the production grammar rule it corresponds to. We leverage these representations to build syntactic features for each word in a sentence, computed as the sum of all the constituents on the path between a word and a task-specific node in the tree, e.g. the target predicate for SRL. Our approach improves state-of-the-art results on the TI and SRL of \texttildelow$1\%$ and \texttildelow3.5\% points, respectively (+2.5$\%$ additional points are gained with BERT as input), when tested on FrameNet 1.5, while yielding comparable results on the CoNLL05 dataset to other syntax-aware systems.
\end{abstract}

\section{Introduction}
\label{sec:intro}
In this paper we focus on the problem of integrating syntactic features in a neural architecture for the Frame-Semantic parsing \cite{Das:2014:FP:2645242.2645244} process. Frame-semantic parsing is the task of extracting full semantic frame structures from text, as  defined by   Frame Semantics theory \cite{Fillmore:85frames}. 
From a theoretical perspective, Frame-Semantic parsing can be decomposed into three sub-tasks: 1) Target Identification (TI) -- identifying target words acting as lexical units; 2)  Frame Identification (FI) -- disambiguating each target into a possible frame; and 3) Semantic Role Labeling (SRL) -- extracting all the possible frame elements for a given frame.

Early neural approaches have focused in this regard on the integration of features extracted from dependency trees, both for the FI and SRL tasks \cite{hermann-etal-2014-semantic, kshirsagar-etal-2015-frame, DBLP:journals/corr/SwayamdiptaTDS17}, with positive results. Amongst all, SRL is the task that has received more attention when investigating methods for injecting syntax into neural models, mostly due to the strict correlation between syntax and argument structures \cite{punyakanok08:importance}.  Several solutions have been proposed, setting new baselines over general Frame-semantic parsing and specific SRL corpora. These include the use of dependency path embeddings \cite{roth-lapata:16:neural}, the application of Graph Convolutional Networks (GCNs) to learn representations of the dependency graphs \cite{marcheggiani-titov:17gcn}, or restricting the set of candidate arguments using pruning algorithms \cite{he18:syntax}. Multi-task learning has been also applied, either directly supervising attention to learn dependency parsing \cite{strubell-etal-2018-linguistically} for both TI and SRL, or to implicitly bias learned encoded representations when jointly training a simplified syntactic dependency parser \cite{swayamdipta-etal-2018-syntactic}, or a semantic dependency parser for both FI and SRL \cite{peng-etal-2018-learning}.

Although effective, these approaches have focused on exploiting syntactic dependencies rather than  constituency information, partly because dependencies are more suited to be encoded as features or learned through attention mechanisms.
Semantic roles are technically provided over syntactic constituents, which directly cast argument boundaries over word sequences. This is demonstrated also by earlier work on SRL, which relied on constituency derived features \cite{Gildea:2002:ALS:643092.643093, xue-palmer04:calibrating, punyakanok08:importance}. It follows that using constituency information should be beneficial, especially because reconstructing argument boundaries through dependencies would require an unbounded number of hops among words, making the problem hard to model in neural architectures \cite{marcheggiani19:axiv}.
Following this idea, two recent approaches have attempted to rely on such constituency information to improve SRL performance. \citet{wang-etal-2019-best} use linearised representations of constituency trees in different learning settings, either by extracting salient features, by multi-task learning, or by combining both approaches in an auto-encoding fashion. \citet{marcheggiani19:axiv}, instead, train a  GCN with the SRL objective to learn constituent representations, which are then infused into words through the same GCN via the message-passing operation \cite{scarselli09:gnn}.

In this paper, we foster the same idea of relying on constituency information for every sub-task of Frame-semantic parsing, namely TI, FI, and SRL. We train a GCN to learn specific constituency representations, which are used in turn to compute syntactic paths between constituency nodes. 
Our approach is similar to that of \citet{marcheggiani19:axiv}, although it significantly differs in: i) the initialisation and topology of the underlying graph; ii) the lower number of required parameters; and iii) the way that syntactic information is infused in every word representation, i.e. computing node-to-node syntactic paths. We show that our approach improves the state-of-the-art over the main Frame-semantic parsing benchmark, i.e.\ the FrameNet corpus \cite{baker-etal-1998-berkeley}, on the single TI and SRL tasks, and on FI in a joint-learning setting. Moreover, we demonstrate the generality of the approach by testing the same network on the CoNLL 2005 dataset \cite{carreras-marquez-2005-introduction}.

\section{Learning Constituent Representations with Graph Convolutional Networks}
\label{sec:gcn}

In this work, we take inspiration from the seminal work on SRL by \citet{Gildea:2002:ALS:643092.643093} to design constituent-based features. In their work, \citeauthor{Gildea:2002:ALS:643092.643093} found discriminant features for a given constituent of a tree to be the path between the target predicate and the constituent itself. Based on the same concept, our idea is to use the path connecting each token with a task-specific reference node as a feature in a sequence labelling setting. For example, following Figure \ref{fig:rachel}, the path $\uparrow$\textsc{vbd}$\uparrow$\textsc{vp}$\downarrow$\textsc{np}$\downarrow$\textsc{jj} between the predicate \textit{had} and the word \textit{little} may provide information to discriminate \textit{little} as the beginning of a potential argument span in SRL.

\begin{figure}[!t]
    \centering
    \includegraphics[width=.7\linewidth]{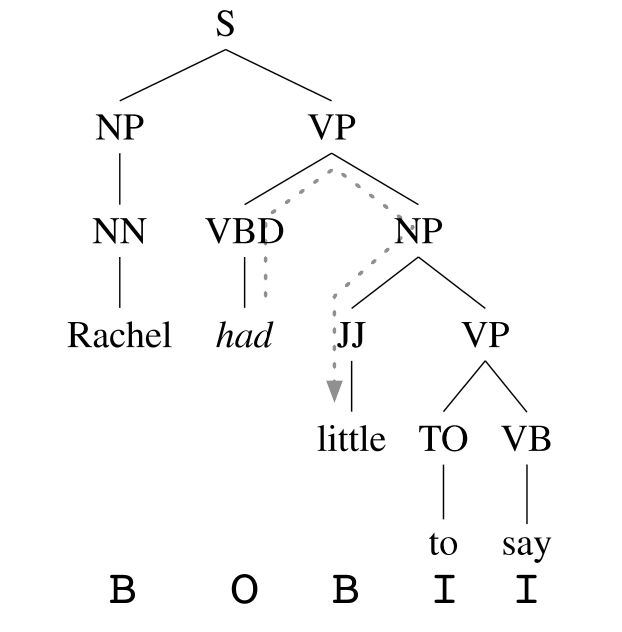}
    \caption{Constituency path between the target predicate \textit{had} and the word \textit{little}. At the bottom, the IOB tagging of the corresponding argument spans.}
    \label{fig:rachel}
\end{figure}

In order to encode a full constituent path as a feature, node representations have to be learned to properly enrich token encodings inside the network. To this end, we train a GCN \cite{Kipf:2016tc}, which operates over a graph representing the constituency tree, learning node (constituent) representations through convolutional steps. According to the message-passing approach, the representation of a constituent is updated with all the representations of the neighbouring nodes at each step. Formally, at the $(l+1)$-th step, the network evaluates the representation
\begin{equation*}
H^{(l+1)} = LN(\sigma(AH^{(l)}W^{(l)} + b^{(l)})),
\end{equation*}
where $LN$ is a layer normalisation, $\sigma$ is the ReLU activation function, $A \in [0,1]^{N \times N}$ is the adjacency matrix of the graph corresponding to a tree, $H^{(l)} \in \mathds{R}^{N \times D}$ is the matrix of activations of the $l$-th layer, while $W^{(l)} \in \mathds{R}^{D \times E}$ and $b^{(l)} \in \mathds{R}^{D \times 1}$ are layer-specific trainable weight matrix and bias. $N$ is the number of nodes in the tree, while $D$ and $E$ are respectively the input and output dimensionality of the $l$-th layer. $H^{(0)} = X$ correspond to the matrix of input representations. We initialise $X$ with trainable embeddings associated with each constituent type, which are learned together with the other representations. The GCN is trained using signal from the task-specific objective function.

\begin{figure}[!t]
    \centering
    \includegraphics[width=\linewidth]{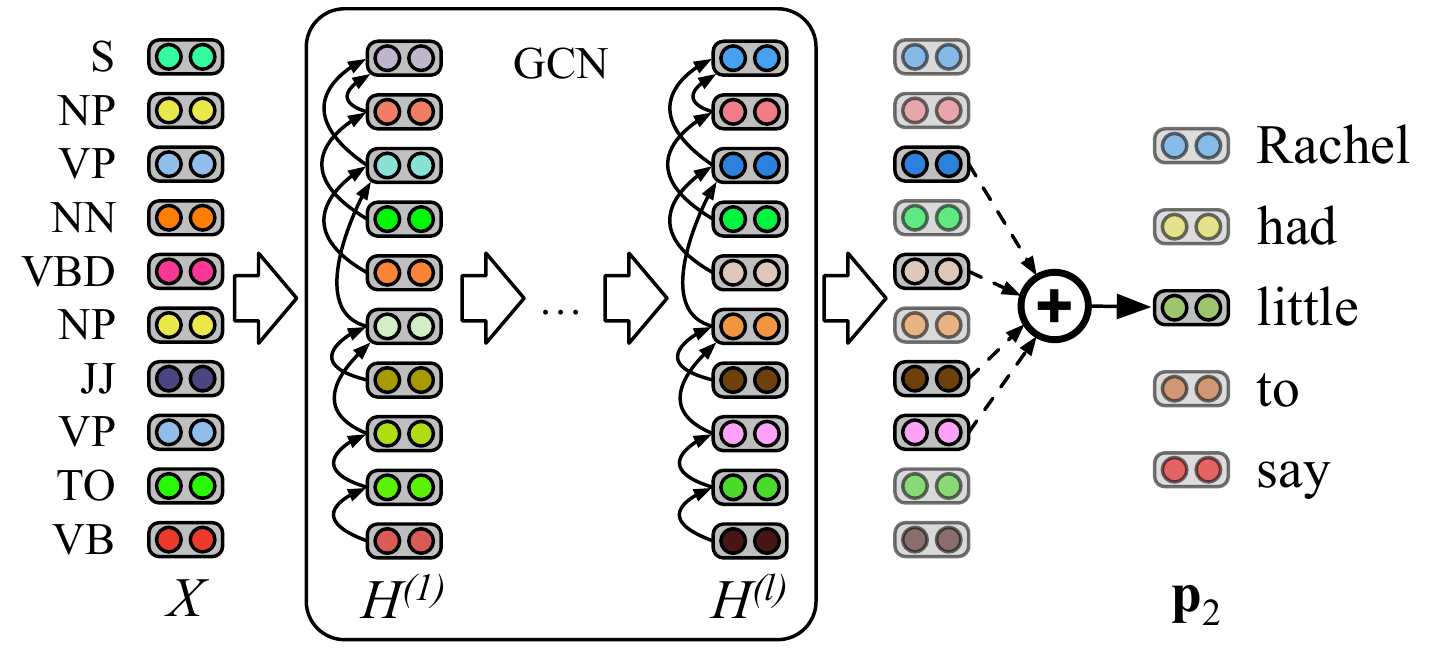}
    \caption{Process of learning  constituent encodings and evaluating path features $\mathbf{p}_2$ from the node \textit{had} over the tree in Figure \ref{fig:rachel}. Detail of feature evaluation for the word \textit{little}.}
    \label{fig:gcn}
\end{figure}

Each father-children relation of a constituency tree represents the production rule of a constituency grammar. We propose to rely on this structural relationship to better profile node representations in our network. To this end, we arrange the graph representing the tree, encoded through $A$, 
only with edges directed
from children to father, so that each constituent is specifically profiled according to the production rule it corresponds to. Self-loops are also added to feed each constituent with its own representation.
Given the constituent encodings $c_1,...,c_N$ composing the rows of the output matrix $H^{(l)}$, 
a path feature $p_{i,j}$ between the nodes $i$ and $j$ is evaluated as
\begin{equation}
    p_{i,j} = \sum_{k \in \mathcal{P}_{i,j}}  c_k,
    \label{eq:paths}
\end{equation}
where $\mathcal{P}_{i,j}$ is set of indexes of nodes connecting the shortest path between $i$ and $j$. Figure \ref{fig:gcn} summarises the whole process of learning constituent representations and evaluating path features $p_{i,j}$.

\section{Integrating Constituency Features in Neural Frame-Semantic Parsing}
\label{sec:architecture}

The process of extracting frame-semantic structures from text typically involves three tasks. Target words evoking frames, i.e. possible \textit{lexical units}, are identified in a sentence during the Target Identification (TI) task. Frame Identification (FI) aims to classify each lexical unit $l$ into a possible frame $f$. For each retrieved $l$ and $f$, SRL is performed to identify and classify the set of related arguments.
In this paper, we want to assess whether our approach of encoding constituency information is beneficial for each of these sub-tasks. 

\begin{figure}[!t]
    \centering
    \includegraphics[width=\linewidth]{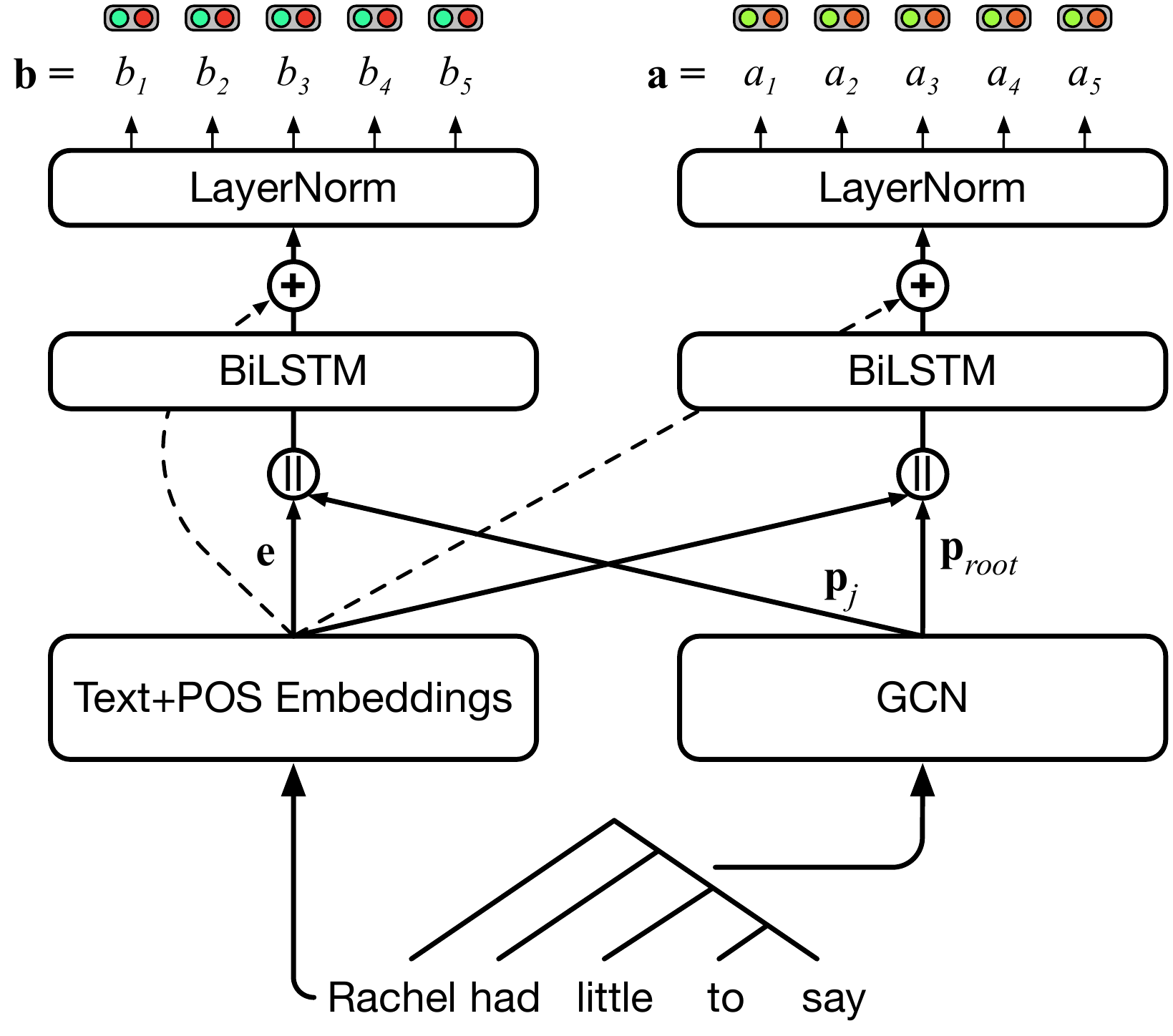}
    \caption{The two encoding backbones. The textual embeddings are fed with the token sequence, while the GCN receives the corresponding syntactic tree as input.}
    \label{fig:encoding}
\end{figure}

In order to pair textual information with the constituency path encodings out of our GCN, we evaluate two encoded sequences $\mathbf{a} = a_1,...,a_n$ and $\mathbf{b} = b_1,...,b_n$  through the encoding block showed in Figure \ref{fig:encoding}, which computes:
 \begin{align}
     \nonumber
     \mathbf{a} = LN(BiLSTM(\mathbf{e}\oplus\mathbf{p}_{root}) + \mathbf{e}), \\
     \mathbf{b} = LN(BiLSTM(\mathbf{e}\oplus\mathbf{p}_l) + \mathbf{e}),
     \label{eq:encodings}
 \end{align}
where $LN$ is a normalization layer, $BiLSTM$ is a bi-directional LSTM encoder \cite{Schuster:1997:BRN:2198065.2205129} and $\mathbf{e} = e_1,...,e_n$ is the sequence of input textual embeddings of a sentence. Residual connections are applied here to $\mathbf{e}$ to allow the gradient to flow when fine-tuning them. The sequences $\mathbf{p_{root}}=p_{1,{root}},...,p_{n,{root}}$ and $\mathbf{p_l}=p_{1,l},...,p_{n,l}$ are sequences of token-wise constituency path features, evaluated as in Equation \ref{eq:paths}, where the reference node $j$ has been set as the root of the tree and a predicate target word corresponding to the lexical unit $l$.\footnote{We use the first word for multi-token targets.} To be consistent with (\ref{eq:paths}), when using the $p_{i,j}$ notation, the positional index $i$ of a word in the input text will correspond to its node index in the tree.\footnote{Since we do not consider the word level in the tree, we create a correspondence between POS-tags and words in term of indexes.} Therefore the encoded sequences $\mathbf{a}$ and $\mathbf{b}$ contain the root-centred path features and the predicate-centered path features, respectively. The TI and FI tasks rely only on the first one, whereas SRL makes use of both the representations, as explained in the following.

\subsection{Target Identification}
\label{subsec:ti}
The TI task is shaped as a sequence labelling task using the IOBC labelling scheme, where the label C is used to deal with discontinued spans. 
Given the input sequence $\mathbf{x} = x_1,...,x_n$, a Conditional Random Field (CRF) layer \cite{Lafferty:2001:CRF:645530.655813} is applied to obtain the sequence of possible target spans $\mathbf{t} = t_1,...,t_n$, with $t_i \in \{\text{\texttt{B-Lu}, \texttt{I-Lu}, \texttt{C-Lu}, \texttt{O}}\}$. For example, the sentence ``\textit{I\textbackslash\texttt{O}} \textit{tried\textbackslash\texttt{B-LU} setting\textbackslash\texttt{B-LU} things\textbackslash\texttt{O} down\textbackslash\texttt{C-LU}}'' has two lexical units: \textit{tried} and \textit{setting down}. The CRF thus models the conditional probability\footnote{From now on, we drop the condition on the learned parameters in the probability for the sake of readability.}
\begin{equation}
\nonumber
P(\mathbf{t} | \mathbf{x}) = \frac{1}{Z}\text{exp}(\sum_{i=1}^n E_{x_i,t_i} + \sum_{i=0}^n T_{t_i,t_{i+1}})
\end{equation}
where emission scores $E_{x_i,t_i}$ are computed from the syntactic-augmented encodings $a_1,...,a_n$, and transitions $T_{t_i,t_{i+1}}$ are learned during training. The most likely label sequence is computed through Viterbi decoding at test time. Given a batch of $D$ sentences, the network is trained by minimising the loss $\mathcal{L}_{TI}= -\frac{1}{D} \sum_{i = 1}^{D} \text{log} P(\mathbf{t}_i | \mathbf{x}_i)$.

\subsection{Frame Identification}
\label{subsec:fi}
Frame Identification is treated as a multi-classification problem over the targets, i.e. lexical units, identified during the TI. We model the probability of a lexical unit $l$ evoking a frame $f$ as 
\begin{equation*}
P(f|l) \propto \text{exp}(W^{(3)} \: \sigma (W^{(2)} \; \sigma(W^{(1)} \; t^l))),
\end{equation*}
where $\sigma$ is the Leaky-ReLU activation function \cite{Maas13rectifiernonlinearities}, and $t^l$ is an encoded representation of $l$ defined as 
\begin{equation}
t^l = \sum_{m \in \mathcal{T}^l} a_m.
\label{fig:tl}
\end{equation}
with $\mathcal{T}^l$ the set of word indexes of the target span for the lexical unit $l$, and $a_m$ are encodings from the sequence in Equation \ref{eq:encodings}. The FI is trained by minimising the average cross-entropy loss 
\begin{equation*}
    \mathcal{L}_{FI} =  - \frac{1}{D} \sum_{i = 1}^{D} 
    \frac{1}{|\mathcal{T}_i|}  \sum_{l \in \mathcal{T}_i} \text{log} P(f|l),
\end{equation*}
where $\mathcal{T}_i$ is the set of gold targets for the $i$-th example.
A mask obtained from the FrameNet ontology is applied to penalise frames that are not evoked by the identified lexical unit.

\subsection{Semantic Role Labeling}
\label{subsec:srl}
Semantic Role Labeling can be operationally decomposed into Argument Identification (AI) and Argument Classification (AC). 
SRL has been recently tackled  with two main approaches. On the one hand, AI is bypassed by computing all the possible sentence subsequences, and then classifying each subsequence against the set $\mathcal{A}^f \cup {\varnothing}$ of possible frame elements for a given frame $f \in \mathcal{F}$, where $\varnothing$ is the no-frame-element label \cite{yang-mitchell-2017-joint}. This approach may lead to limiting the maximum length of the computed subsequences, leaving out some wide-covering arguments. On the other hand, AI and AC have been tackled together as an IOB-based sequence labelling task, where the label space has been defined as $\{\{\texttt{B}, \texttt{I}\}\times \mathcal{A}^{\mathcal{F}} \} \cup \{\texttt{O}\}$ \cite{strubell-etal-2018-linguistically}. When it comes to datasets like FrameNet, where $|\mathcal{A}^{\mathcal{F}}| = \text{725}$, the number of parameters increases drastically, especially when bilinear matrix operations are applied to score tokens against predicates, where the sole bilinear layer can reach up to \texttildelow95M parameters.

\begin{figure}[!t]
    \centering
    \includegraphics[width=\linewidth]{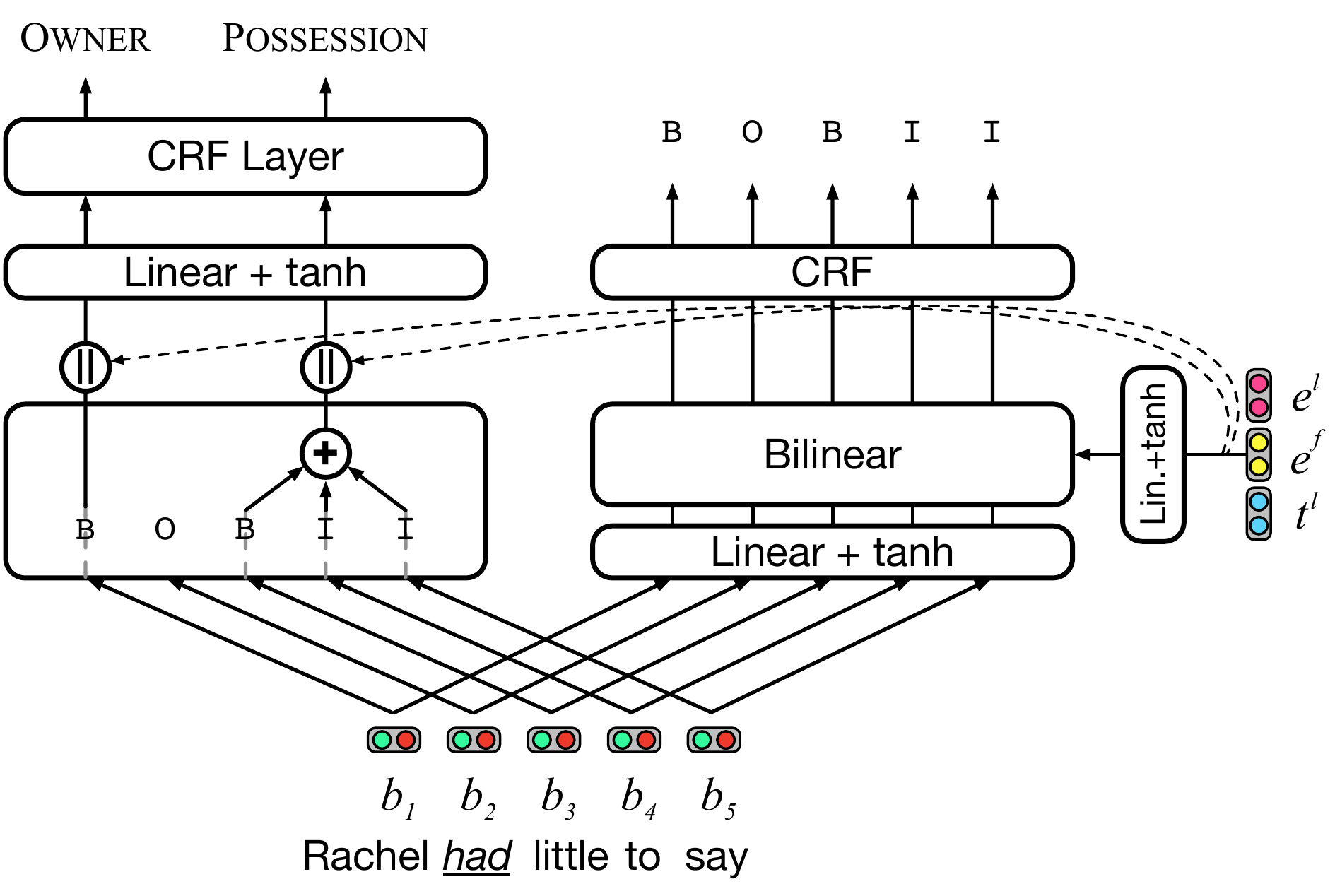}
    \caption{AI and AC sub-networks on the right and on the left, respectively.}
    \label{fig:srl}
\end{figure}

To overcome this issue, 
we treat the AI and AC as two separate tasks, making the number of parameters drop from 95M to \texttildelow3M\footnote{These numbers indeed are proportional to the dimensionality of the inputs. However, their ratio would be the same. }. We use a hierarchy of two CRFs, one for each task (see Fig. \ref{fig:srl}). The first extracts the argument span from a sentence representation using a bilinear operation. The second, instead, operates over the resulting sequence of spans, assigning a frame element label to each of them. This second CRF helps in contextualising frame elements with each other. 

\paragraph{Argument Identification (AI)} For a frame $f$ and related lexical unit $l$, the AI task is modelled as the conditional probability $P(\mathbf{s}|\mathbf{x},l, f)$, where $\mathbf{s} = s_1,...,s_n$ is a sequence of label in the IOB2 notation, 
i.e. $s_i \in \{\text{\texttt{B}, \texttt{I}, \texttt{O}}\}$, denoting the argument spans. Similarly as in Section \ref{subsec:ti}, we employ a CRF to model the conditional probability, with the difference that emission scores are computed through a bilinear operation between tokens and predicate representations, in order to capture interactions between each token $x_i$ and the current pair $(l,f)$. Given $l$, $f$ and the encoding of the $i$-th token $b_i$ from Eq. \ref{eq:encodings}, we compute per-frame IOB2 token emissions through the bilinear operation
\begin{gather}
    pr^{(l,f)\top} \; U \; pb_i, \; \:  pr^{(l,f)} = tanh(V^{(1)}z^{(l,f)}), \\
    \nonumber 
    pb_i = tanh(V^{(2)} b_i),
    \label{eq:predicate}
\end{gather}
with $z^{(l,f)} = e^l \oplus t^l \oplus e^f$, and $V^{(1)},V^{(2)}$ are trainable weight matrices. Here, $e^l$ and $e^f$ are lexical unit and frame learned embeddings, while $t^l$ is the target representation evaluated in Equation \ref{fig:tl}.

\paragraph{Argument Classification (AC)} Given a sequence of argument spans $\mathbf{s}^f$\footnote{We drop the $l$ subscript for the sake of readability.}
for a frame and lexical unit $(l, f)$ from the AI step, the AC is modelled as the conditional probability $P(\mathbf{a}|\mathbf{x}, l, f, \mathbf{s}^f)$, where $\mathbf{a} = a_1,...,a_u$ is a sequence of frame element labels, with each $a_w \in \mathcal{A}^f$ 
being an frame element label associated with the $w$-th argument span from $\mathbf{s}^f$. Single vector representations $r_w$ for each span are computed as
\begin{equation*}
    r_w = \sum_{m \in \mathcal{S}^f_w} b_m,
\end{equation*}
where $\mathcal{S}^f_w$ is the subset of word indexes of the $w$-th argument span, extracted from $\mathbf{s}^f$. 
Each vector in the sequence $r_1,...,r_u$ is then concatenated with the predicate representation $z^{(l,f)}$ from Equation \ref{eq:predicate}, and projected through $q_w = tanh(Y r_w \oplus z^{(l,f)})$. $P(\mathbf{a}|\mathbf{x}, l, f, \mathbf{s}^f)$ is evaluated with a CRF Layer which computes emission probabilities over the inputs $q_1,...,.q_u$. The SRL is trained by optimising the loss $\mathcal{L}_{SRL}$ given by 
\begin{align}
    - \frac{1}{D} \sum_{i = 1}^{D} 
    \frac{1}{|\mathcal{F}_i|}  \sum_{(l,f,\mathbf{s}) \in \mathcal{F}_i} \Big[ 
    \nonumber & \text{log} P(\mathbf{s}|\mathbf{x}_i, l, f) + \\
    \nonumber & \text{log} P(\mathbf{a}|\mathbf{x}_i, l, f, \mathbf{s}^f) \Big],
\end{align}
where $\mathcal{F}_i = \{\langle l_1, f_1, \mathbf{s}_1 \rangle,...,\langle l_q, f_q, \mathbf{s}_q \rangle\}$ is the set of gold frame annotations for the $i$-th example. At inference time, Viterbi decoding is applied for both the AI and AC tasks to evaluate the most probable tag and frame element label sequences. A theory-based mask extracted from the FrameNet ontology is applied to the emission matrix to penalise frame element labels that are not associated with the current frame. 

\section{Experimental Evaluation}

This section reports the evaluation procedure and results obtained by the proposed model on the different tasks with respect to existing approaches.

\subsection{Experimental Setup}
\label{sec:data}

Constituency path features in our architectures are validated firstly on FrameNet 1.5 (FN15) \cite{baker-etal-1998-berkeley}, and on CoNLL-2005 (CN05) \cite{carreras-marquez-2005-introduction} to verify the generality of the approach. 
For both datasets, we apply the original data split 
as in \citet{Das:2014:FP:2645242.2645244} for FN15, and as in \citet{carreras-marquez-2005-introduction} on CN05. 
Results are obtained through the official evaluation of the SemEval07 shared task for frame-semantic parsing \cite{baker-etal-2007-semeval} (FN15) and the CoNLL-2005 shared task for Semantic Role Labeling (CN05).\footnote{\url{https://www.cs.upc.edu/~srlconll/soft.html}}

The model is implemented using PyTorch \cite{paszke2017automatic} and AllenNLP \cite{Gardner2017AllenNLP}, and trained on RTX 2080Ti GPUs.
The best hyperparameters are obtained tuning the model on the development set, via grid search using early stopping.
We concatenate POS tag embeddings to input text embeddings, obtained using spaCy \cite{spacy2} for FN15, or directly from gold annotations for CN05.
Constituent trees are obtained through the Berkeley Neural Parser \cite{Kitaev-2018-SelfAttentive}, using the English pre-trained model on FN15; for CN05 we generate the syntactic trees of the development and test sets using the model obtained on the CN05 training set. For the GCN, only $l=2$ layers have been used, to make each constituent take representations only from its children and grandchildren. We use the Adam optimizer \cite{Adam} with weight decay fix (BERTAdam), with an initial learning rate of $2\times 10^{-5}$, which is decreased once no improvements are observed for 5 epochs. We regularise the model using dropout on encoders and feed-forward layers, while L2 regularisation is used on the CRFs' transition matrices and bilinear weight matrix.\footnote{All the code, experimental setup, data and parameter configurations for all the experiments available at \url{www.Anon.}} 

We test different configurations of  textual embeddings (OURS-*), disabling or enabling the GCN features (*-GCN). For example, in \architecturename-ELMo we use ELMo \cite{peters2018} as input layer, without syntax features. We also include a joint learning setting (\architecturename-BERT-GCN-JL), where the three tasks are jointly optimised by summing the losses ($\mathcal{L} = \mathcal{L}_{\text{TI}} + \mathcal{L}_{\text{FI}} + \mathcal{L}_{\text{SRL}}$) and BERT \cite{devlin-etal-2019-bert} is used as input layer.

\subsection{FrameNet 1.5 Results}

In this section, we report the evaluation on FN15 for each tasks presented in Section~\ref{sec:architecture}.


\paragraph{Target Identification}

\begin{table}[]
	\centering
	\footnotesize
	\begin{tabular}{lccc}
	\toprule
		{Model} & P & R & F1\\
		\midrule
		{\citet{Das:2014:FP:2645242.2645244}} & {$37.5$} & {$57.5$} & {$45.4$}\\
		{\citet{DBLP:journals/corr/SwayamdiptaTDS17}} & {$-$} & {$-$} & {$73.2$}\\
		{\textbf{\architecturename{}}-ELMo} & {$69.84$} & {$\textit{78.88}$} & {$74.08$}\\
		{\textbf{\architecturename{}}-ELMo-GCN} & {$\textit{70.08}$} & {$78.72$} & {$\textit{74.15}$}\\
		\midrule
		{\textbf{\architecturename{}}-BERT} & {$71.71$} & {$78.51$} & {$74.96$}\\
		{\textbf{\architecturename{}}-BERT-GCN} & {$70.66$} & {$\textbf{84.12}$} & {$\textbf{76.80}$}\\
		{\textbf{\architecturename{}}-BERT-GCN-JL} & {$\textbf{72.59}$} & {$79.43$} & {$75.86$}\\
		\bottomrule
	\end{tabular}
	\caption{Precision, Recall and F1 of the Target Identification task.
	\label{tab:ti_fn}}
\end{table}
Table~\ref{tab:ti_fn} shows the results of TI. In addition to neural state-of-the-art systems, we also report the rule-based baseline computed on the FN15 test set following\citet{Das:2014:FP:2645242.2645244}.
The proposed architecture outperforms existing approaches in the literature, improving the state-of-the-art model in \citet{DBLP:journals/corr/SwayamdiptaTDS17} by \texttildelow$1\%$ of F1 on comparable inputs, and an additional $2.65\%$ points are gained when using BERT.
The syntax encoding contribution seems to be evident across the different input layers, with a more consistent improvement ($1.84\%$) on BERT.
Joint learning here acts as a regulariser, improving the Precision with respect to the single task setting. The model exposes good generalisation capabilities, correctly identifying $46.47\%$ of the 402 out-of-vocabulary targets contained in the test set. 

However, the final performance of TI is not easy to assess, mostly due to the well-known issue of the \fnversion{} test set of missing a considerable amount of unannotated targets \cite{yang-mitchell-2017-joint}.

\paragraph{Frame Identification}

\begin{table}[]
	\centering
	\footnotesize
	\begin{tabular}{lccc}
	\toprule
		{Model} & {} & {} & Acc\\
		\midrule
		{\citet{Das:2014:FP:2645242.2645244}} & {} & {} & {$83.6$}\\
		{\citet{hermann-etal-2014-semantic}} & {} & {} & {$88.4$}\\
		{\citet{hartmann-etal-2017-domain}} & {} & {} & {$87.6$}\\
		{\citet{yang-mitchell-2017-joint}} & {} & {} & {$88.2$}\\
		{\citet{peng-etal-2018-learning}} & {} & {} & {$89.9$}\\
		{\textbf{\architecturename{}}-ELMo} & {} & {} & {$88.89$}\\
		{\textbf{\architecturename{}}-ELMo-GCN} & {} & {} & {$88.82$}\\
		\midrule
		{\textbf{\architecturename{}}-BERT} & {} & {} & {$89.90$}\\
		{\textbf{\architecturename{}}-BERT-GCN} & {} & {} & {$89.83$}\\
		{\textbf{\architecturename{}}-BERT-GCN-JL} & {} & {} & {$\textbf{90.10}$}\\
		\bottomrule
	\end{tabular}
	\caption{Frame Identification results using gold targets, in terms of Accuracy.
	\label{tab:fi}}
\end{table}

In Table~\ref{tab:fi} we report the results of FI using gold targets.
All the configurations are comparable with results in literature, while \architecturename-BERT-GCN-JL outperforms the state-of-the-art model from \citet{peng-etal-2018-learning} by $0.2$, suggesting that jointly learning the whole frame-semantic parsing process provides shared representations that result beneficial for the FI. However, this result is in line with \citet{peng-etal-2018-learning}, where multi-task learning is used to exploit the contribution of a purely syntactic task.
Overall, our features do not seem to provide a direct contribution for FI when trained in a single-task fashion.

\paragraph{Semantic Role Labeling}
\begin{table}[]
	\centering
	\footnotesize
	\begin{tabular}{lccc}
	\toprule
		{Model} & P & R & F1\\
		\midrule
		{\citet{Das:2014:FP:2645242.2645244}$^{\dagger}$} & {$65.6$} & {$53.8$} & {$59.1$}\\
		{\citet{kshirsagar-etal-2015-frame}$^{\dagger}$} & {$66.0$} & {$60.4$} & {$63.1$}\\
		{\citet{yang-mitchell-2017-joint}$^{\dagger}$} & {$70.2$} & {$60.2$} & {$65.5$}\\
		{\citet{swayamdipta-etal-2018-syntactic}} & {$69.2$} & {$69.0$} & {$69.1$}\\
		{\citet{marcheggiani19:axiv}} & {$69.8$} & {$68.8$} & {$69.3$}\\
		{\textbf{\architecturename{}}-ELMo} & {$63.89$} & {$67.36$} & {$65.58$}\\
		{\textbf{\architecturename{}}-ELMo-GCN} & {$\textit{72.02}$} & {$\textit{73.70}$} & {$\textit{72.85}$}\\
		\midrule
		{\textbf{\architecturename{}}-BERT} & {$71.19$} & {$74.26$} & {$72.69$}\\
		{\textbf{\architecturename{}}-BERT-GCN} & {$74.23$} & {$\textbf{76.94}$} & {$\textbf{75.56}$}\\
		{\textbf{\architecturename{}}-BERT-GCN-JL} & {$\textbf{74.56}$} & {$74.43$} & {$74.50$}\\
		\bottomrule
	\end{tabular}
	\caption{Semantic Role Labeling results with gold targets and frames.
	\label{tab:srl_gold}}
\end{table}

SRL results are shown in Table~\ref{tab:srl_gold}.  
The proposed model obtains state-of-the-art results, outperforming all the existing approaches on SRL with gold targets and frames, with an improvement of $3.55\%$ over \citet{marcheggiani19:axiv} in comparable settings, and gaining additional $2.71$ points when applying BERT.
SRL is the task where syntax encodings provide the largest contribution, especially when paired with the ELMo contextual embeddings. In this setting, GCNs allow to gain up to $7.27\%$ of F1. 
This behaviour is mainly due to the nature of the SRL task. In fact, as already reported in Section \ref{sec:intro}, 
argument boundaries are defined by the syntactic constituents spanning sub-trees. Such relationships seem to be well captured by our constituency path features.
However, when using BERT as input layer, such a large improvement is not observed (i.e. ``only'' $2.87\%$). This is due to the deeper contextual representations provided by BERT, whose Transformer-based architecture already captures meaningful representations of the sentence's syntactic structure. In fact, it BERT's ability of specialising some of its attention heads to recover short-range syntactic dependencies has been observed  \cite{htut2019attention,jawahar-etal-2019-bert,hewitt2019structural}; here we show that pairing this information to explicit constituent encoding allows to produce richer representations.
Jointly learning the tasks does not result in learning better shared representations for SRL, although the higher Precision suggests a regularisation benefit of such a learning pattern.

\subsection{Impact of syntax: ablation study}
We finally run an ablation study on FN15, in order to show the clear contribution of constituency path features when changing different input textual embeddings. For this analysis, we select SRL, because it seems to be the task benefitting the most from the syntactic features.
\begin{table}[]
	\centering
	\footnotesize
	\begin{tabular}{lccc}
	\toprule
		 & \multicolumn{2}{c}{SRL}\\
		{Input} &  \textsc{No-Gcn} & \textsc{Gcn} & gain \\
		\midrule
		{GloVe} & {$49.13$} & {$68.23$} & $+19.1$\\
		{ELMo} & {$65.58$} & {$72.85$} & $+7.27$\\
		{BERT} & {$72.69$} & {$75.56$} & $+2.87$\\
		\bottomrule
	\end{tabular}
	\caption{Ablation on the use of constituency features with different input layers for the SRL on FN15.
	\label{tab:ablation}}
\end{table}
Table \ref{tab:ablation} reports the results of this ablation.
It is interesting to see how the contribution of constituency feature encodings decreases as richer and more complex word embeddings are introduced. In fact, contextual embeddings like ELMo and BERT already encode some sort of dependency relations. However, results show that our constituency features complement such syntactic information. 
The contribution of syntax is strongly evident especially when using GloVe non-contextual embeddings, bringing the model $1$ point from the previous state-of-the-art \cite{marcheggiani19:axiv}, which uses contextual embeddings, with an absolute improvement of $19.1\%$. 

\subsection{CoNLL-2005 Results}


In the following, we report the experiment on CN05, performed to
assess the ability of the proposed model to scale to similar problems. 
We use the hyperparameters obtained on FN15, and tune the best epoch on the CN05 development set. 

\paragraph{Target Identification/Predicate Detection}

\begin{table}[]
	\centering
	\footnotesize
	\begin{tabular}{lccc}
	\toprule
		{Model} & P & R & F1\\
		\midrule
		\textit{WSJ Test} &  &  & \\
		\midrule
		{\citet{he-etal-2017-deep}} & {$94.5$} & {$98.5$} & {$96.4$}\\
		{\citet{strubell-etal-2018-linguistically}} & {$98.9$} & {$97.9$} & {$98.4$}\\
		\midrule
		{\textbf{\architecturename{}}-GloVe} & {$98.39$} & {$99.01$} & {$98.70$}\\
		{\textbf{\architecturename{}}-GloVe-GCN} & {$98.64$} & {$98.88$} & {$98.76$}\\
		\midrule
		\midrule
		\textit{Brown Test} &  &  & \\
		\midrule
		{\citet{he-etal-2017-deep}} & {$89.3$} & {$95.7$} & {$92.4$}\\
		{\citet{strubell-etal-2018-linguistically}} & {$95.5$} & {$91.9$} & {$93.7$}\\
		\midrule
		{\textbf{\architecturename{}}-GloVe} & {$90.15$} & {$95.65$} & {$92.82$}\\
		{\textbf{\architecturename{}}-GloVe-GCN} & {$90.55$} & {$92.91$} & {$91.71$}\\
		\bottomrule
	\end{tabular}
	\caption{Predicate detection results on the CN05 dataset. 
	\label{tab:ti_conll}}
\end{table}

Table \ref{tab:ti_conll} shows the results of Target Identification (or Predicate Detection). For fairness, we report results of our model without contextual embeddings, since other approaches in the literature rely on these input layers. The proposed model obtains state-of-the-art results for in-domain examples, whereas it seems to suffer from overfitting when applied to out-of-domain data.

\paragraph{Semantic Role Labeling}

We report the results of our approach for SRL on CoNLL-2005.
For such analysis, we use BERT as input layer, best-performing on FN15.
\begin{table}[]
	\centering
	\footnotesize
	\begin{tabular}{lccc}
	\toprule
		{Model} & P & R & F1\\
		\midrule
		\textit{WSJ Test} &  &  & \\
		\midrule
		{\citet{he-etal-2018-jointly}} & {$84.2$} & {$83.7$} & {$83.9$}\\
		{\citet{tan2018deep}} & {$84.5$} & {$85.2$} & {$84.8$}\\
		{\citet{li2019dependency}} & {$87.9$} & {$87.5$} & {$87.7$}\\
		{\citet{strubell-etal-2018-linguistically}$^\dagger$} & {$86.02$} & {$86.05$} & {$86.04$}\\
		{\citet{ouchi-etal-2018-span}} & {$88.2$} & {$87.0$} & {$87.6$}\\
		{\citet{wang-etal-2019-best}} & {$-$} & {$-$} & {$88.2 $}\\
		{\citet{marcheggiani19:axiv}} & {$87.7$} & {$88.1$} & {$87.9$}\\
		\midrule
		{\textbf{\architecturename{}}-BERT} & {$87.01$} & {$87.36$} & {$87.18$}\\
		{\textbf{\architecturename{}}-BERT-GCN} & {$87.46$} & {$87.87$} & {$87.66$}\\
		\midrule
		\midrule
		\textit{Brown Test} &  &  & \\
		\midrule
		{\citet{he-etal-2018-jointly}} & {$74.2$} & {$73.1$} & {$73.7$}\\
		{\citet{tan2018deep}} & {$73.5$} & {$74.6$} & {$74.1$}\\
		{\citet{li2019dependency}} & {$80.6$} & {$80.4$} & {$80.5$}\\
		{\citet{strubell-etal-2018-linguistically}$^\dagger$} & {$76.7$} & {$76.4$} & {$76.5$}\\
		{\citet{ouchi-etal-2018-span}} & {$76.0$} & {$70.4$} & {$73.1$}\\
		{\citet{wang-etal-2019-best}} & {$-$} & {$-$} & {$79.3$}\\
		{\citet{marcheggiani19:axiv}} & {$80.5$} & {$80.7$} & {$80.6$}\\
		\midrule
		{\textbf{\architecturename{}}-BERT} & {$78.76$} & {$79.56$} & {$79.16$}\\
		{\textbf{\architecturename{}}-BERT-GCN} & {$80.44$} & {$81.21$} & {$80.82$}\\
		\bottomrule
	\end{tabular}
	\caption{SRL results with gold predicates on the CN05 dataset. $\dagger$ models do not use contextual embeddings.
	\label{tab:srl_conll}}
\end{table}
Table \ref{tab:srl_conll} shows the results of such experiments. Although our model does not overcome state-of-the-art results, it shows its competitiveness.
Both on the in-domain (WSJ) and out-domain (Brown) test sets, syntactic encodings provide an important contribution, consistently improving the performance of the network. The smaller gap between the syntactic and non-syntactic one with respect to results over FN15 may be partially due to the different (and higher) complexity of the CN05 constituency trees. In fact, when looking at the constituent vocabulary, we observe that the small amount of available constituents in FN15 (74) is not even comparable to the vocabulary size of CN05 constituents (1170). 


\section{Related Work}

\paragraph{Graph Neural Networks to encode syntax} The structural information conveyed by syntactic trees has been encoded in neural models through Graph Neural Networks (GNNs) \cite{scarselli09:gnn} in several text processing applications. The Syntactic GCN framework has been applied to incorporate syntactic \cite{bastings17:nmt} and semantic \cite{marcheggiani:18syntgcn} structures by applying GCN to Neural Machine Translation. On the same task, \citet{beck18:graph} build a graph-to-sequence encoder using a Gated GNN \cite{li16:gated}, which accounts for the whole dependency structures by adding dependency labels as additional nodes to the graph. Syntax encoding has been investigated also in the context of the Event Extraction task, that is the extraction of specified classes of events from texts. \citet{nguyen18:event} apply the Syntactic GCN framework to inject dependency information to characterise event detection. \citet{liu18:jointly}, instead, leverage syntactic dependency shortcuts to build local sub-graph representations of each node, encoded through an Attention-based GCN. Graph convolutions have  also been applied to learn syntactic and semantic embeddings \cite{vashishth19:incorporating} in an unsupervised fashion, which are shown to provide advantages when integrated in other tasks. Finally, \citet{Li18:clinic} propose a Segment Graph Convolutional and Recurrent NN, which operates over word embedding and syntactic dependencies, to classify relations from clinical notes.

\paragraph{Syntax-aware neural models for Frame-semantic parsing}
Identifying solutions to model syntax in neural models to improve Frame-semantic parsing related tasks has been a main concern in the community in the last five years. These have ranged from manual feature engineering \cite{hermann-etal-2014-semantic, kshirsagar-etal-2015-frame, DBLP:journals/corr/SwayamdiptaTDS17}, to learning dependency path embeddings \cite{roth-lapata:16:neural}, to encoding syntactic dependency structures via GCNs \cite{marcheggiani-titov:17gcn}. Learning syntactic \cite{strubell-etal-2018-linguistically} or semantic \cite{peng-etal-2018-learning} dependency parsing as auxiliary tasks in a multi-task learning settings has found successful application, especially when the sub-task outputs were directly injected as features. 

Only two approaches have tackled the problem of integrating constituency information. \citet{wang-etal-2019-best} propose to linearise the constituency tree using the model from \citet{gomez-rodriguez-vilares18:constituent}, and apply a combination of learning techniques to improve sequence labelling performance. \citet{marcheggiani19:axiv}, instead, apply a GCN over the constituency tree to learn constituent boundaries to be injected in word sequence representations. In this work, we also apply a GCN, although our primary aim is to directly learn constituent representations. Whereas \citeauthor{marcheggiani19:axiv} start from an empty tree updated only with word embedding features, we initialise it with constituent specific embeddings. Moreover, we rely on a directed graph to profile constituent representations as the grammar rules they correspond to. This configuration allows us to learn a more compact model, with a single matrix of parameters for each layer, against the six weight matrices required by their approach.\footnote{Please refer to \citet{marcheggiani19:axiv} for details.} Finally, we inject syntactic representations into words by explicitly computing constituency path features.

\section{Conclusions}
In this work, we investigated the integration of structural information from a constituent tree in a neural model for Frame-semantic parsing. Constituent representations are learned through a GCN, 
and used to build constituency path features to be added to every word representation in a sequence.
We tested our approach on all the Frame-semantic parsing sub-tasks, namely Target Identification, Frame Identification, and Semantic Role Labeling, showing that such features contribute mainly on the TI and the SRL tasks. 

Constituency path features can be applied
to other sequence labelling based tasks, e.g. Named-Entity Recognition. Moreover, other modifications of GCNs have to be tested in this same framework, e.g. to assess whether Attention-based GCN may learn more refined constituent representations. Finally, these representations may be used in a node-classification approach, inspired by seminal works \cite{Gildea:2002:ALS:643092.643093}, in an attempt to move away from the well-used sequence labelling model of   recent years.

\bibliographystyle{acl_natbib}

\bibliography{emnlp2020.bib}


\end{document}


\appendix
\section{Model hyperparameters}
This appendix reports the hyperparameter configuration used in our experiments.
The model is implemented using PyTorch and AllenNLP, and trained on RTX 2080Ti GPUs. The best hyperparameters are obtained tuning the model on the development set, via grid search using early stopping.
\subsection{Input Embeddings}
The textual input is represented through a textual embeddings, i.e. one among GLoVE, ELMo, or BERT, and POS-tags embeddings.

\paragraph{GLoVe embeddings parameters}
\begin{itemize}
    \item Embedding dimension: 300
\end{itemize}
\paragraph{ELMo embeddings parameters}
\begin{itemize}
    \item Embedding dimension: 1152
    \item Layer normalisation enabled
\end{itemize}
\paragraph{BERT embeddings parameters}
\begin{itemize}
    \item Embedding dimension: 768
\end{itemize}
\paragraph{POS-tags embeddings parameters}
\begin{itemize}
    \item Embedding dimension: 20
\end{itemize}
\subsection{Graph Convolutional Network}
The GCN used to learn constituent encodings has the following topology:
\begin{itemize}
    \item Number of convolutional layers $l$:2
    \item Input constituent embeddings dimension: 128
    \item Hidden dimension: 128
    \item Output dimension: 128
    \item Activation function: ReLU
    \item Dropout: 20\%
\end{itemize}
\subsection{Shared Encoding Layer}
The shared encoding layer is implemented through two Bidirectional LSTMs, one used by all the tasks, and one used only during SRL. Both the LSTMs share the same topology, described in the following:
\begin{itemize}
    \item Hidden dimension: 394
    \item Number of layers: 2
    \item Dropout: 20\%
\end{itemize}
\subsection{Target Identification}
The Target Identification is realised with a Conditional Random Field Layer decoder with the following parameters: 
\begin{itemize}
    \item Constrained CRF decoding enabled
    \item Regulariser: L2 regularisation on transition matrix
\end{itemize}
\subsection{Frame Identification}
The Frame Identification multi-classification is performed with a 3-layer Feed Forward Network with the following parameters:
\begin{itemize}
    \item Hidden dimensions: 788, 788, 512
    \item Activation function: LeakyRelu
    \item Dropout: 20\%
\end{itemize}
\subsection{Semantic Role Labeling}
The sub-network performing Semantic Role Labeling is composed by two modules resolving Argument Identification and Argument Classification.
\subsubsection{Argument Identification}
The AI uses a Linear layer to project the predicates, and a Bilinear operation to score tokens against predicates. Bilinear logits are used in a CRF as emission probabilities.
\paragraph{Linear projection layer}
\begin{itemize}
    \item Hidden dimension: 128
    \item Dropout: 20\%
    \item Activation: tanh
\end{itemize}
\paragraph{Bilinear matrix attention}
\begin{itemize}
    \item Dimension 1: 256
    \item Dimension 2: 256
    \item Regulariser: L2 on the weigh matrix
\end{itemize}
\paragraph{Conditional Random Field}
\begin{itemize}
    \item Hidden dimension 1: 256
\end{itemize}
\subsection{Argument Classification}
The AC uses a Linear layer to project concatenation of argument spans and predicate representations, and score the argument sequence using a CRF Layer.
\paragraph{Linear projection layer}
\begin{itemize}
    \item Hidden dimension: 256
    \item Dropout: 20\%
    \item Activation: tanh
\end{itemize}
\paragraph{CRL Layer}
\begin{itemize}
    \item Constrained CRF decoding enabled
    \item Regulariser: L2 regularisation on transition matrix
\end{itemize}

\subsection{Training}
The specific BERT version of Adam is used in training as an optimiser, with parameters
\begin{itemize}
    \item Epsilon: $1 \times 10^{-8}$
    \item Learning rate: $2 \times 10^{-5}$
    \item Epochs before decrease: 5
\end{itemize}

A learning rate scheduler reducing the learning rate on plateau is used, in $min$ mode, i.e. the scheduler monitors the decrease of the inspected quantity. Parameters of the scheduler are the following:
\begin{itemize}
    \item Patrience: 5
    \item Reducing factor: 0.5
    \item New optimum threshold: $1 \times 10^{-4}$
    \item Threshold mode: Absolute
    \item Epsilon: $1 \times 10^{-8}$
\end{itemize}

Early stopping monitors the target performance depending on the task or the learning modality. F1 is used for TI and SRL, Accuracy for FI, and a combination of FI Accuracy and SRL F1 is used when in joint-learning mode. Other training parameters are:
\begin{itemize}
    \item Max number of epochs: 200
    \item Gradient clipping: 10
    \item Rescale gradient norm: 5
    \item Patience: 100
\end{itemize}

Total number of parameters by task:
\begin{itemize}
    \item Target Identification: XXM
    \item Frame Identification: YYM
    \item Semantic Role Labeling: ZZM
    \item Joint Learning (TI+FI+SRL): WWM
\end{itemize}

Wherever no parameters are reported, we assume default values from the AllenNLP implementation.